\documentclass[10pt, a4paper]{article}

\usepackage[final]{lrec2026} 

\usepackage{booktabs}
\usepackage{multirow}

\title{Irish-BLiMP: A Linguistic Benchmark for Evaluating Human and Language Model Performance in a Low-Resource Setting}

\name{
\large \bfseries
Josh McGiff\textsuperscript{1,*}, Khanh-Tung Tran\textsuperscript{2,*}, William Mulcahy, Dáibhidh Ó Luinín,\\
\large \bfseries Jake Dalzell, Róisín Ní Bhroin, Adam Burke, Barry O'Sullivan\textsuperscript{2},\\
\large \bfseries Hoang D. Nguyen\textsuperscript{2}, Nikola S. Nikolov\textsuperscript{1}
}

\address{\textsuperscript{1}University of Limerick, Ireland\\
         \textsuperscript{2}University College Cork, Ireland\\[1ex]
         Corresponding author: josh.mcgiff@ul.ie
         }

\abstract{
We present Irish-BLiMP (Irish Benchmark of Linguistic Minimal Pairs), the first dataset and framework designed for fine-grained evaluation of linguistic competence in the Irish language, an endangered language. Drawing on a variety of linguistic literature and grammar reference works, we manually constructed and reviewed 1020 minimal pairs across a taxonomy of 11 linguistic features, through a team of fluent Irish speakers. We evaluate both existing Large Language Models (LLMs) and fluent human participants on their syntactic knowledge of Irish. Our findings show that humans outperform all models across all linguistic features, achieving 16.6\% higher accuracy on average. Moreover, a substantial performance gap of 18.1\% persists between open- and closed-source LLMs, with even the strongest model (gpt-5) reaching only 73.5\% accuracy compared to 90.1\% by human. Interestingly, human participants and models struggle on different aspects of Irish grammar, thus highlighting a difference in representation learned by the models. Overall, Irish-BLiMP provides the first systematic framework for evaluating the grammatical competence of LLMs in Irish and offers a valuable benchmark for advancing research on linguistic understanding in low-resource languages.
\newline  
}

\begin{document}

\maketitleabstract

\section{Introduction}
The rise of the Transformer architecture \cite{vaswani2017attention} has empowered technological advancements across the field of natural language processing (NLP). In terms of NLP, transformers have been instrumental in improving the state–of-the-art for language modelling tasks like sentiment analysis \cite{McGiff2024Bridging}, neural translation \cite{nllb2024scaling} and dialogue generation \cite{zhang2019dialogpt}. The widespread popularity of large language model (LLM) services like ChatGPT and Gemini can be attributed to their capacity to boost productivity and automate complex systems \cite{filippucci2024impact}. 

Although the impact of transformer-powered services on English and other majority language-based domains is an active research area, low-resource language communities are often an afterthought. Low-resource language communities suffer from the challenges associated with data scarcity and the limited availability of computational resources \cite{mcgiff2025overcoming}. The increasing integration of transformer-powered services in public and private sectors \cite{dper_ai_guidelines2025, mckinsey_state_of_ai2025}, despite their limited compatibility with many minority languages, is exacerbating existing forms of technological language inequality. The existing technological biases where English-language content is used by roughly half of the websites on the internet~\cite{commoncrawl}, is perpetuated in the training of flagship LLMs. Therefore, more research is required to assess existing LLMs on their understanding of linguistic features in low-resource languages like Irish.

Some work has assessed the linguistic capabilities of natural languages by composing a specialised dataset focused on the evaluation of specific syntactic features. Research has previously focused on majority languages like English \cite{warstadt2020blimp} and Chinese \cite{xiang2021climp}, with some studies extending to other widely spoken European languages \cite{suijkerbuijk2025blimp, barbini2025blimp} and non-European languages \cite{someya2023jblimp}. Until now, there has not been any formal research investigating the linguistic capability of LLMs on Celtic languages. 

As a result, we have developed Irish-BLiMP (Benchmark of Linguistic Minimal Pairs for Irish). We created 1020 Irish-language minimal pairs for targeted syntactic evaluation. These minimal pairs are created and divided into 11 categories based on linguistic literature relating to the Irish language \cite{stenson2019modern,OHanluain1999,AnCaighdean2017}. Some of the 11 macro categories contain sub-categories for finer granularity, with each feature consisting of 10 handcrafted sentences. For example, the verbal morphology category contains various sub-categories that explore each of the verb tenses by standard formation, autonomous form, irregular formations and other related phenomena. Overall, this paper contributes:
\begin{itemize}
\item A novel dataset and an accompanying framework (Irish-BLiMP) for the syntactic evaluation of LLMs on their linguistic understanding of the Irish language.
\item Through extensive experiments, we reveal a substantial performance gap between state-of-the-art open-source and closed-source models. Open-source models perform near the random baseline, while the strongest closed-source model (gpt-5) achieves only 73.5\% accuracy, indicating significant gap in multilingual grammatical generalisation.
\item A human-LLM comparative assessment demonstrates that the benchmark is straightforward for humans (average human accuracy of 90.1\%), yet challenging for LLMs. This contrast suggests that current models rely heavily on surface pattern recognition rather than an internalised understanding of Irish grammatical rules.
\end{itemize}

Irish-BLiMP aims to provide Irish-language LLM researchers with a novel dataset and an accompanying framework for the syntactic evaluation of their models. Unlike typical language model evaluation metrics like BLEU \cite{papineni2002bleu} and COMET \cite{rei2020comet}, our discriminative evaluation framework assesses existing open-source and proprietary models based on a taxonomy of grammatical features specific to the Irish language. This evaluation framework aims to enable researchers to assess future generative language models based on Irish-language specific features. 

\section{Related Work}
The linguistic evaluation of language models for low-resource languages remains under-researched. Recent studies developing language models for LRLs often rely on automatically calculated metrics like BLEU, COMET and GLUE/SuperGLUE \cite{wang2018glue, wang2019superglue, mcgiff2025overcoming}. While these studies highlight a high-level view of the performance of their models, these metrics do not provide granular insights into the syntactic, linguistic understanding of the models. In other words, automatically generated metrics cannot indicate the specific aspects of language that generative models accurately or inaccurately represent. This context is necessary to inform processes for improving language modelling capacity for LRLs. 

Previous studies researching language modelling for the Irish language often select from a mixture of evaluation approaches such as using automatically generated metrics and performing human-based evaluation  \cite{barry2021gabert, tran2024irish, tran2024uccix, mcgiff2025semiadaptsemiloraefficientdomain}. These approaches suffer from the aforementioned challenge of identifying Irish language-specific linguistic phenomena that the language models struggle with or perform well on. Although some syntactic insights can be found through expert human evaluation of downstream tasks like neural translation \cite{clifford2025gaeilge}, a wider assessment of linguistic features is required to highlight and report generalisations across a model’s overall understanding of a language like Irish. 


An early approach for evaluating the linguistic capabilities of language models introduced the CoLA dataset \cite{warstadt2019cola} by mapping examples from linguistics literature to binary acceptability labels. However, CoLA is limited due to its use of a classifier, and is improved in a subsequent study that introduces BLiMP \cite{warstadt2020blimp}. BLiMP’s approach uses linguist-crafted grammar templates to automatically generate 67 datasets of minimal pairs. A minimal pair consists of an acceptable and an unacceptable sentence that differ only with respect to a single targeted linguistic phenomenon. However, the reliance of BLiMP and other approaches like BLiMP-IT \cite{barbini2025blimp} and CLIMP \cite{xiang2021climp} on templates for automatically generating minimal pairs could produce trivial and inaccurate sentences or omit edge-case linguistic features that linguistic experts or literature would cover. Moreover, the frequent use of a limited, predefined vocabulary for automatic generation increases the risk of incomplete coverage of a language model’s capacity to capture the full range of syntactic phenomena \cite{barbini2025blimp, xiang2021climp}. Additionally, given that approaches for morphologically rich languages like Dutch \cite{suijkerbuijk2025blimp} opt for a larger number of linguistic feature categories with semi-synthetic minimal pairs, it could indicate that Irish requires more complex templates to create non-trivial minimal pairs in comparison with BLiMP. 

JBLiMP \cite{someya2023jblimp} addresses the pitfalls of automatically generating minimal pairs for Japanese by sourcing 2,323 acceptability judgements from journal articles based on Japanese syntax. Similar to BLiMP, JBLiMP categorised the dataset by type, phenomenon, and paradigm. Similarly, BLiMP-NL \cite{suijkerbuijk2025blimp} creates minimal pairs by handcrafting 10 sentence pairs per paradigm. However, BLiMP-NL utilises a language model (GPT-3.5 Turbo) to generate additional minimal pairs from the handcrafted sentence pairs. Although the synthetic sentences were reportedly checked manually and corrected, the generation hinges on the chosen model’s linguistic competence for the language as is. For low-resource languages where language models currently have limited understanding of them, this would be more prone to inaccuracies \cite{mcgiff2025overcoming}.

In terms of dataset validation, methods differ between human evaluation using crowd-sourced and expert participants. BLiMP validation participants only validate a fraction of examples per paradigm \cite{warstadt2020blimp}. In contrast, each minimal pair in JBLiMP is validated by individual participants \cite{someya2023jblimp}. CLiMP enlists at least two native speakers \cite{xiang2021climp} and BLiMP-IT employs at least two linguists \cite{barbini2025blimp}. Alternatively, BLiMP-NL \cite{suijkerbuijk2025blimp} represents the most comprehensive validation in the BLiMP family with each paradigm being tested by at least 30 native Dutch speakers who provided graded acceptability judgements on a 7-point scale and participated in self-paced reading experiments. TurBLiMP, by comparison, only reports acceptability judgement experiments using a 7-point Likert scale with 30 native Turkish speakers \cite{bacsar2025turblimp}. Overall,  studies differ in the number of participants for human validation based on their fluency and expertise in the area of linguistics for the language. 

Although a BLiMP-based approach covering 101 languages has been developed \cite{jumelet2025multiblimp}, some languages are represented with very few minimal pairs. For example, results are reported for Irish evaluation across various models with 28 minimal pairs. Given that the study only covers basic linguistic phenomena for Irish, the results and insights lack sufficient granularity to meaningfully evaluate model competence in the language. Therefore the MultiBLiMP approach is rather limited in its validity for Irish and other underrepresented languages.

\section{Irish-BLiMP}
\subsection{Data}
We compiled a list of acceptability judgements by consulting an array of grammar books for the Irish language \cite{stenson2019modern, OHanluain1999,AnCaighdean2017}. Based on these sources, we devised a taxonomy of linguistic phenomena and manually constructed ten minimal pairs for each feature. Our labour-intensive approach of manually creating sentence pairs contrasts with the original BLiMP approach \cite{warstadt2020blimp}. Additionally, we do not use LLMs to generate minimal pairs.  

\begin{figure}[htbp]
        \centering
        \includegraphics[width=0.5\textwidth]{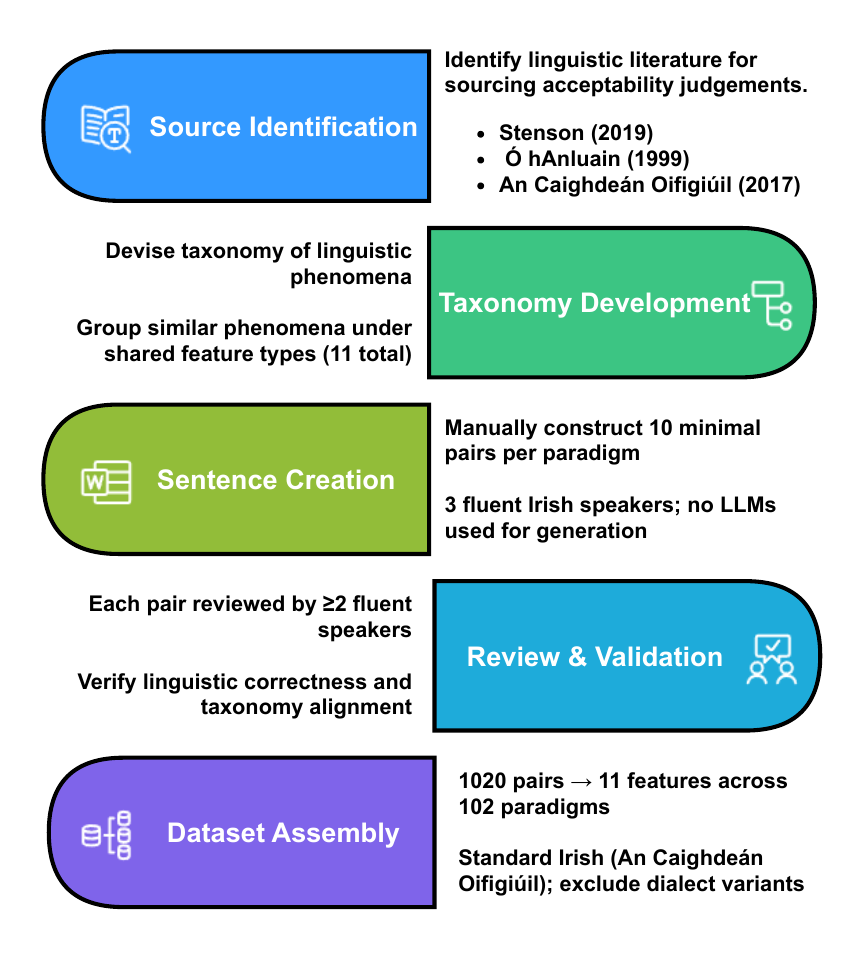}
        \caption{Overview of the data creation process for the Irish minimal pairs dataset, from source identification to final dataset assembly.}
        \label{fig:datadiagram}
    \end{figure}

We enlisted three fluent Irish speakers to produce 1020 minimal pairs spanning 11 linguistic features. Although the final taxonomy includes 11 features aggregated from various linguistic sources, these could be further subdivided to separate closely related phenomena, such as clause structure and word order. Each sentence pair was reviewed by at least two fluent participants to ensure linguistic correctness. The participants created both acceptable and unacceptable sentence pairs in alignment with the taxonomy of features provided. It is worth noting that the acceptability judgements are based on the standard form of Irish, known as An Caighdeán Oifigiúil and they do not apply to dialect-specific features \cite{AnCaighdean2017}. We report the dataset statistics in Table~\ref{tab:dataset_structure} and describe the data creation process in Figure~\ref{fig:datadiagram}.

\begin{table}[t]
\centering
\resizebox{1.0\linewidth}{!}{
\begin{tabular}{p{1.5cm} p{4cm} c c}
\hline
\textbf{Type} & \textbf{Phenomenon} & \multirow{2}{*}{\shortstack{\textbf{\#Para-}\\\textbf{digms}}} & \textbf{Total} \\
\\
\hline

\multirow{3}{1.5cm}{\textbf{Verbal morphology}} 
& Tense and mood & 28 & 310 \\
& Verb agreement suffixes & 2 &  \\
& Negation and polarity & 1 &  \\
\hline

\multirow{4}{1.5cm}{\textbf{Nouns and cases} }
& Gender & 1 & 120 \\
& Case system & 5 &  \\
& Nouns and adjectives & 5 &  \\
& Abstract nouns & 1 &  \\
\hline

\multirow{4}{1.5cm}{\textbf{Adjectives \& Comparison} }
& Comparative and superlative degrees & 1 & 40 \\
& Irregular comparatives & 1 &  \\
& Equative structures & 1 &  \\
& Adjective predicates with the copula & 1 &  \\
\hline

\multirow{6}{1.5cm}{\textbf{Pronouns} }
& Personal pronouns & 1 & 90 \\
& Possessive adjectives & 2 &  \\
& Progressive functions & 1 &  \\
& Prepositional pronouns & 3 &  \\
& Relative pronouns & 1 &  \\
& Pronoun objects & 1 &  \\
\hline

\multirow{3}{1.5cm}{\textbf{Articles \& Determiners} }
& Definite article & 1 & 30 \\
& Article in prepositional contexts & 1 &  \\
& Determiners & 1 &  \\
\hline

\multirow{3}{1.5cm}{\textbf{Numbers} }
& Cardinals and nouns & 1 & 30 \\
& Ordinal numbers & 1 &  \\
& Personal numbers & 1 &  \\
\hline

\textbf{Copula \& bí} 
& Copula uses & 6 & 60 \\
\hline

\multirow{6}{1.5cm}{\textbf{Clause structure \& word order} }
& Basic VSO order & 1 & 170 \\
& Other orders & 1 &  \\
& Progressive sentences & 1 &  \\
& Verbal nouns & 5 &  \\
& Subordinate clauses & 6 &  \\
& Relative clauses & 3 &  \\
\hline

\multirow{5}{1.5cm}{\textbf{Questions \& negations} }
& Yes/No questions & 1 & 50 \\
& Tag questions & 1 &  \\
& Constituent questions & 1 &  \\
& Indirect questions & 1 &  \\
& Answers & 1 &  \\
\hline

\multirow{4}{1.5cm}{\textbf{Discourse \& Sentence types} }
& Reported speech & 1 & 40 \\
& Focus constructions in copula sentences & 1 &  \\
& Cleft questions and answers & 1 &  \\
& Embedded clefts & 1 &  \\
\hline

\multirow{5}{1.5cm}{\textbf{Adverbs \& Modifiers} }
& Manner adverbs & 4 & 80 \\
& Time and place adverbs & 1 &  \\
& Directional adverbs & 1 &  \\
& Evaluative adjectives & 1 &  \\
& Sentential adverbs & 1 &  \\
\hline
\textbf{Total} 
&  & \textbf{102} & \textbf{1020} \\
\hline
\end{tabular}
}
\caption{Overview of dataset structure organised by linguistic \textbf{Type}, \textbf{Phenomenon} and number of \textbf{Paradigms}}
\label{tab:dataset_structure}
\end{table}

\subsection{Linguistic Phenomena Coverage}
We devised a taxonomy of 11 linguistic phenomena based on grammatical features described in Irish linguistic literature and grammar reference works \cite{stenson2019modern, OHanluain1999,AnCaighdean2017}. We echo the BLiMP framework \cite{warstadt2020blimp}, by organising the linguistic phenomena at three hierarchical levels: type, phenomenon and paradigm. Type refers to a broad label encompassing similarly related phenomena. Phenomena denotes specific linguistic features for a given type, such as the case system phenomena as part of the nouns and cases type. A paradigm represents more granular features of a given phenomenon, such as the genitive system as part of the case phenomenon. The following section breaks down the taxonomy by type, phenomena and subsequent paradigms.

\subsubsection{Verbal Morphology}
This category examines tense, mood, and agreement in verbal forms, including habitual tense contrasts, subject agreement suffixes replacing pronouns, and the interaction between negation and emphatic or focus particles.

\noindent\textbf{Example:}
\begin{itemize}
    \item \textbf{Acceptable:} \textit{Baineadh geit aisti ar maidin.} (“She got a fright in the morning.”)
    \item \textbf{Unacceptable:}    \textit{Bhaineadh geit aisti ar maidin.}\\
    The autonomous past tense is used without lenition.
\end{itemize}

\subsubsection{Nouns and Cases:}
This section examines nominal morphology, focusing on gender agreement, case marking across nominative, genitive, dative, and vocative forms, and the interaction of adjectives with case and number distinctions.

\noindent\textbf{Example:}

\begin{itemize}
    \item \textbf{Acceptable:} \textit{Cailín bocht.}\\
    (“Poor girl.”)

    \item \textbf{Unacceptable:}    \textit{Cailín bhocht.}\\
    Adjectives following nominative singular masculine nouns are not lenited.
\end{itemize}

\subsubsection{Adjectives and Comparison:}
This section addresses adjectival morphology and syntax, including comparative and superlative formation, irregular comparison patterns, equative constructions, and the use of adjectives as predicates with the copula.

\noindent\textbf{Example:}

\begin{itemize}
    \item \textbf{Acceptable:} \textit{An duine is ciúine.}\\
    (“The quietest person.”)

    \item \textbf{Unacceptable:} \textit{An duine is ciúin.}\\
    Adjectives ending in a consonant are palatalised with \textit{-e}.
\end{itemize}

\subsubsection{Pronouns}

This section explores pronominal forms and functions, including personal and possessive pronouns, prepositional and relative constructions, pronoun objects, and the use of pronouns in progressive and possessive contexts.

\noindent\textbf{Example:}

\begin{itemize}
    \item \textbf{Acceptable:} \textit{Sin ár n-arán-na.}\\
    (“That’s our bread.”)

    \item \textbf{Unacceptable:} \textit{Sin ár arán-na.}\\
    Plural possessive pronouns prepend \textit{n-} to nouns starting with a vowel.
\end{itemize}

\subsubsection{Articles and Determiners}
This section examines the use of definite articles, their behavior in prepositional contexts involving mutation, and the role of determiners in expressing definiteness, quantity, and specificity.

\noindent\textbf{Example:}

\begin{itemize}
    \item \textbf{Acceptable:} \textit{Ar an lampa.}\\
    (“On the lamp.”)

    \item \textbf{Unacceptable:} \textit{Ar an nlampa.}\\
    \textit{Ar an} causes eclipsis, but \textit{l} cannot be eclipsed.
\end{itemize}

\subsubsection{Numbers}
This section addresses interaction of numerals with nouns, including mutation patterns with cardinals, formation and placement of ordinals, and the distinct use of personal numbers for counting people.

\noindent\textbf{Example:}

\begin{itemize}
    \item \textbf{Acceptable:} \textit{Ochtar fear.}\\
    (“Eight men.”)

    \item \textbf{Unacceptable:} \textit{Ochtar fhear.}\\
    \textit{Ochtar} does not lenite initial consonants.
\end{itemize}

\subsubsection{Copula and Bí}
This section examines functions of the copula and the verb \textit{bí}, including their roles in equational and cleft constructions, relative forms, and contrasts in expressing identity, state, possession, and location.

\noindent\textbf{Example:}

\begin{itemize}
    \item \textbf{Acceptable:} \textit{Is cluichire mé.}\\
    (“I am a gamer.”)

    \item \textbf{Unacceptable:} \textit{Tá mé cluichire.}\\
    The copula is required when the predicate is a noun.
\end{itemize}

\subsubsection{Clause Structure and Word Order}
This section explores clause organization and syntactic patterns, including basic verb–subject–object (VSO) order, marked word orders for focus or emphasis, the use of verbal nouns in progressive and subordinate constructions, and the structure of complement, causal, temporal, and relative clauses.

\noindent\textbf{Example:}

\begin{itemize}
    \item \textbf{Acceptable:} \textit{Labhróidh Liam amárach.}\\
    (“Liam will speak tomorrow.”)

    \item \textbf{Unacceptable:} \textit{Liam labhróidh amárach.}\\
    Irish is a VSO language.
\end{itemize}

\subsubsection{Questions and Negation}
This section examines interrogative and negative structures, including yes/no and tag questions, constituent and indirect questions, and the formation of appropriate response sentences.

\noindent\textbf{Example:}

\begin{itemize}
    \item \textbf{Acceptable:} \textit{An labhraíonn tú Gaeilge?}\\
    (“Do you speak Irish?”)

    \item \textbf{Unacceptable:} \textit{Ar labhraíonn tú Gaeilge?}\\
    Present tense questions use \textit{an}, not \textit{ar}.
\end{itemize}

\subsubsection{Discourse and Sentence Types}
This section explores discourse-level and sentence-type variation, including reported speech, copular focus constructions, cleft questions and answers, and embedded clefts within complex clauses.

\noindent\textbf{Example:}

\begin{itemize}
    \item \textbf{Acceptable:} \textit{Cheap mé go raibh siad deas.}\\
    (“I thought they were nice.”)

    \item \textbf{Unacceptable:} \textit{Cheap mé go bhí siad deas.}\\
    The indirect relative clause uses \textit{go raibh}.
\end{itemize}

\subsubsection{Adverbs and Modifiers}
This section examines formation and function of adverbs, including manner, time, place, directional types, as well as evaluative and sentential adverbs that convey attitude or discourse-level meaning.

\noindent\textbf{Example:}

\begin{itemize}
    \item \textbf{Acceptable:} \textit{Tagann siad anseo go hannamh.}\\
    (“They come here rarely.”)

    \item \textbf{Unacceptable:} \textit{Tagann siad anseo go annamh.}\\
    An \textit{h} is inserted before a vowel following \textit{go}.
\end{itemize}

\section{Experiment Setup}
    \subsection{Evaluation Method}
        Our dataset is intended to evaluate the grammatical knowledge of LLMs across diverse linguistic phenomena in the Irish language. We adopt a multiple-choice evaluation paradigm, inspired by the original BLiMP benchmark (which introduces the minimal pair and asks LMs to prefer the grammatical over the ungrammatical option)~\cite{warstadt2020blimp}. This format is widely used and is compatible with frameworks such as lm-evaluation-harness~\cite{eval-harness}.

       For open-source models, we compute the output probability (log-likelihood) of each choice (A vs. B), then select the option with the higher score as the model’s prediction. For API-based (closed-source) models, we prompt the model with the question and compare its textual reply against the correct label. We enforce determinism and reproducibility by setting the temperature to 0, minimise or disable reasoning effort for reasoning-capable models (gpt-5, gpt-oss-120b, and gpt-oss-20b), and require exact-match of the label (A or B).

    \subsection{Evaluated Models}
        We benchmark prominent state-of-the-art LLMs. This includes leading closed-source models such as gpt-5~\cite{gpt5}, claude-sonnet-4.5~\cite{claude45}, and gemini-2.5-flash-lite~\cite{deepmind_gemini_flash}, known for superior performance across various tasks, as well as open-source models, namely gpt-oss-120b, gpt-oss-20b~\cite{openai2025gptoss120bgptoss20bmodel}, Llama-4-Scout-Instruct~\cite{llama4}, gemma-3-27b-it~\cite{gemma_2025}, and Mistral-Small-3.2-24B-Instruct~\cite{mistral_small}.

        To our knowledge, none of the frontier multilingual models explicitly support Irish. To this end, we also include UCCIX~\cite{tran2024uccix}, a model fine-tuned specifically on Irish, as well as its base model (Llama-2-13B~\cite{touvron2023llama2openfoundation}), to quantify the effect of Irish-specific adaptation.

    \subsection{Human Baseline}
        To ground model performance, we conduct human evaluation with three native Irish speakers, yielding total 3,060 annotations, and average accuracy of 90.09\% (agreement rate: 77.27\%), indicating that the minimal pair task is straightforward for native speakers, confirming that the items are valid and not ambiguous. Moreover, the gap between human and model performance (as shown in the next section) shows that the task remains challenging for current models, prompting further research on LLMs for extremely low-resource languages.

	\subsection{Prompting Strategies.}
		We explore three prompting strategies to examine how context or additional examples helps grammatical competence of LLMs:
        \begin{itemize}
            \item Zero-shot: no exemplars or explanations, comparable to the human experiment.
            \item Few-shot: we create five additional pairs as examples for the LLMs, helping them to understand the task and format requirements.
            \item Grammar-book context: for each paradigm, a brief description of it is provided alongside the current pair. This helps us analysing in-context grammatical understanding capabilities of LLMs, whether they can learn and deduce the grammatically correct option.
        \end{itemize}
    	Among these, only zero-shot prompting is directly comparable to the human evaluation setup, as few-shot and grammar-context prompts give LLMs extra advantages not available to humans. Therefore, zero-shot is our baseline that aligns directly with the human test conditions.

\begin{figure*}[h]
    \centering
    \includegraphics[width=0.95\textwidth]{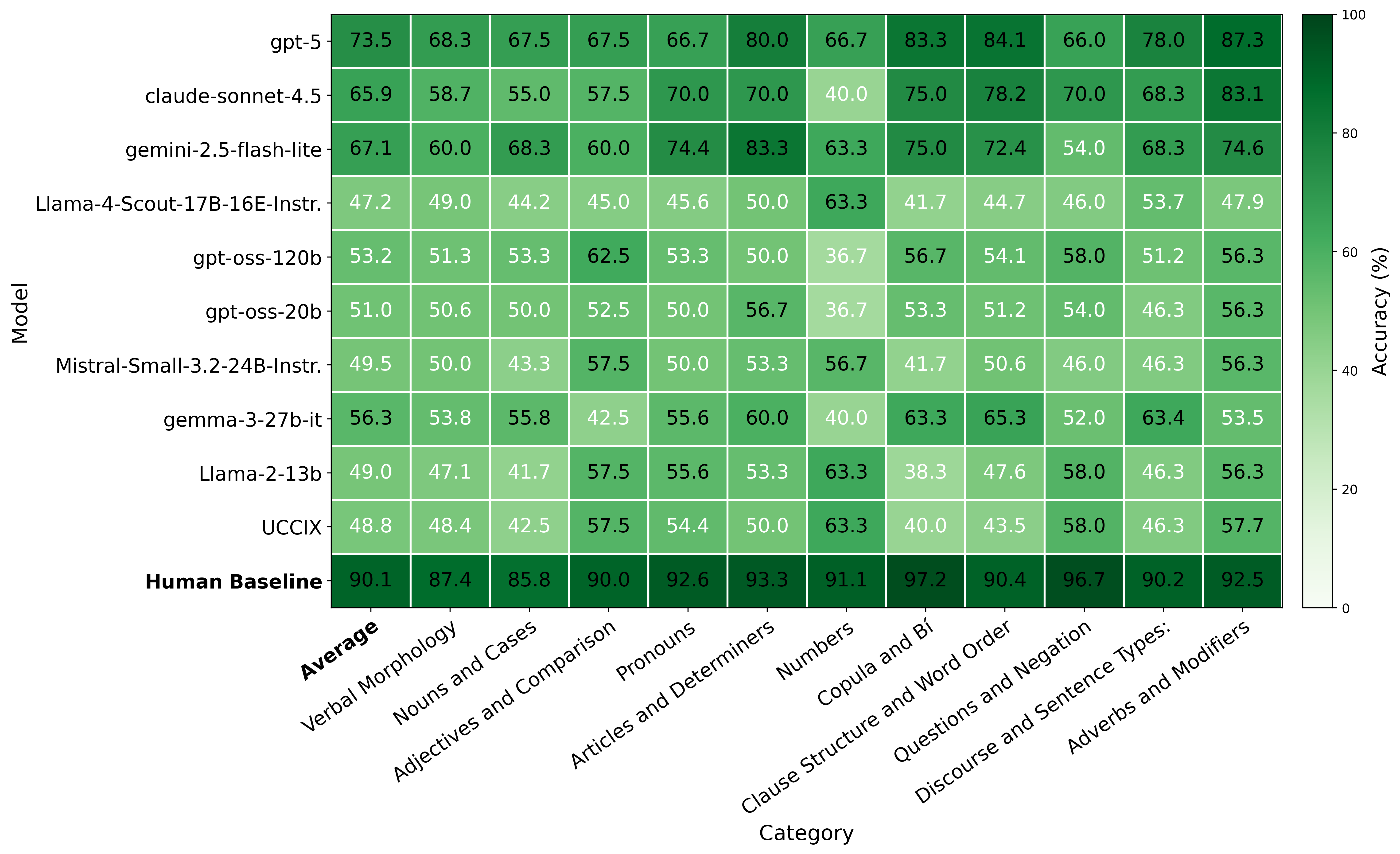}
    \caption{Zero-shot accuracy (\%) by model and category.}
    \label{fig:zeroshot}
\end{figure*}

\begin{figure*}[h]
    \centering
    \includegraphics[width=0.75\textwidth]{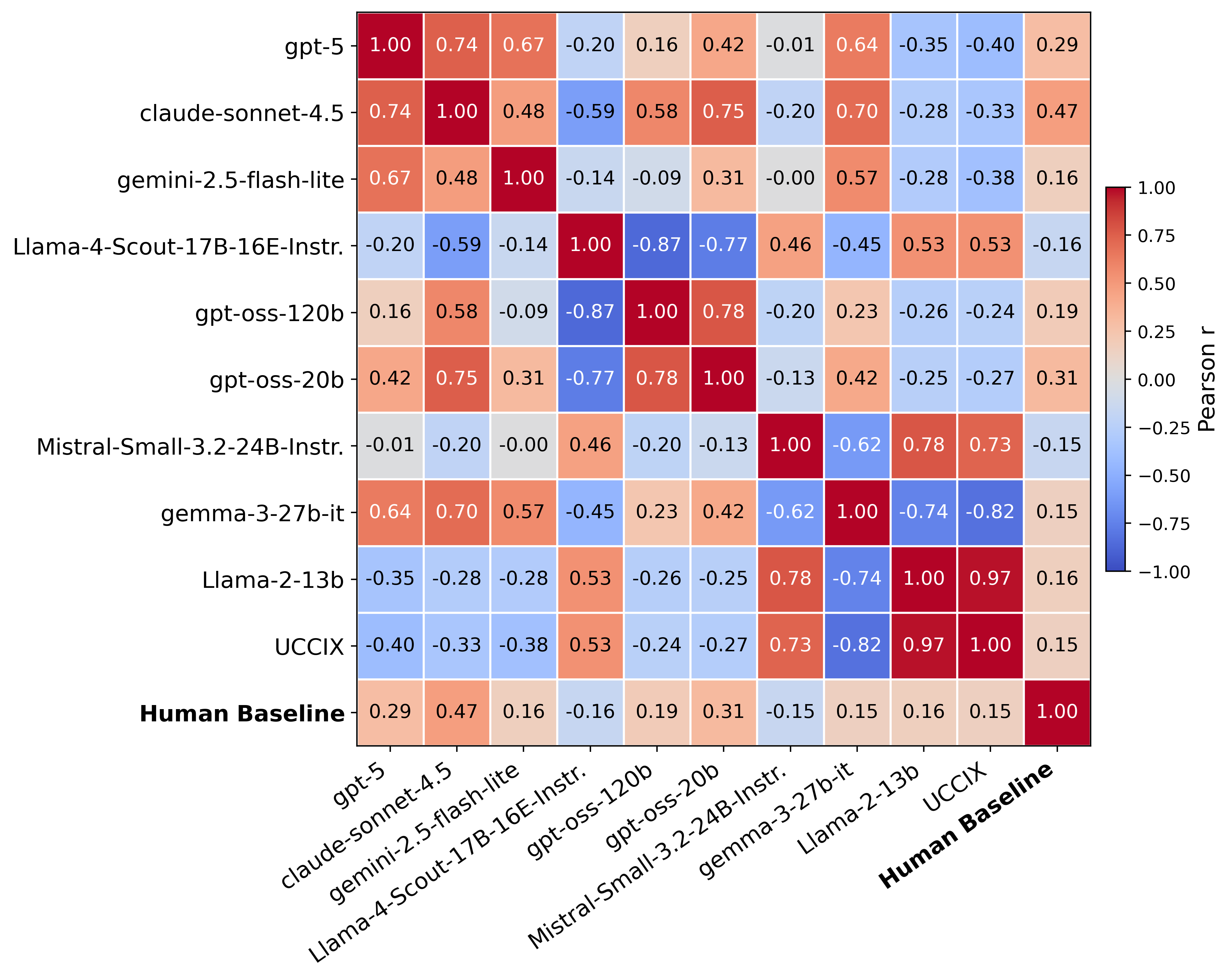}
    \caption{Correlation between models across categories.}
    \label{fig:correlation}
\end{figure*}

\begin{figure*}[h]
    \centering
    \includegraphics[width=0.9\textwidth]{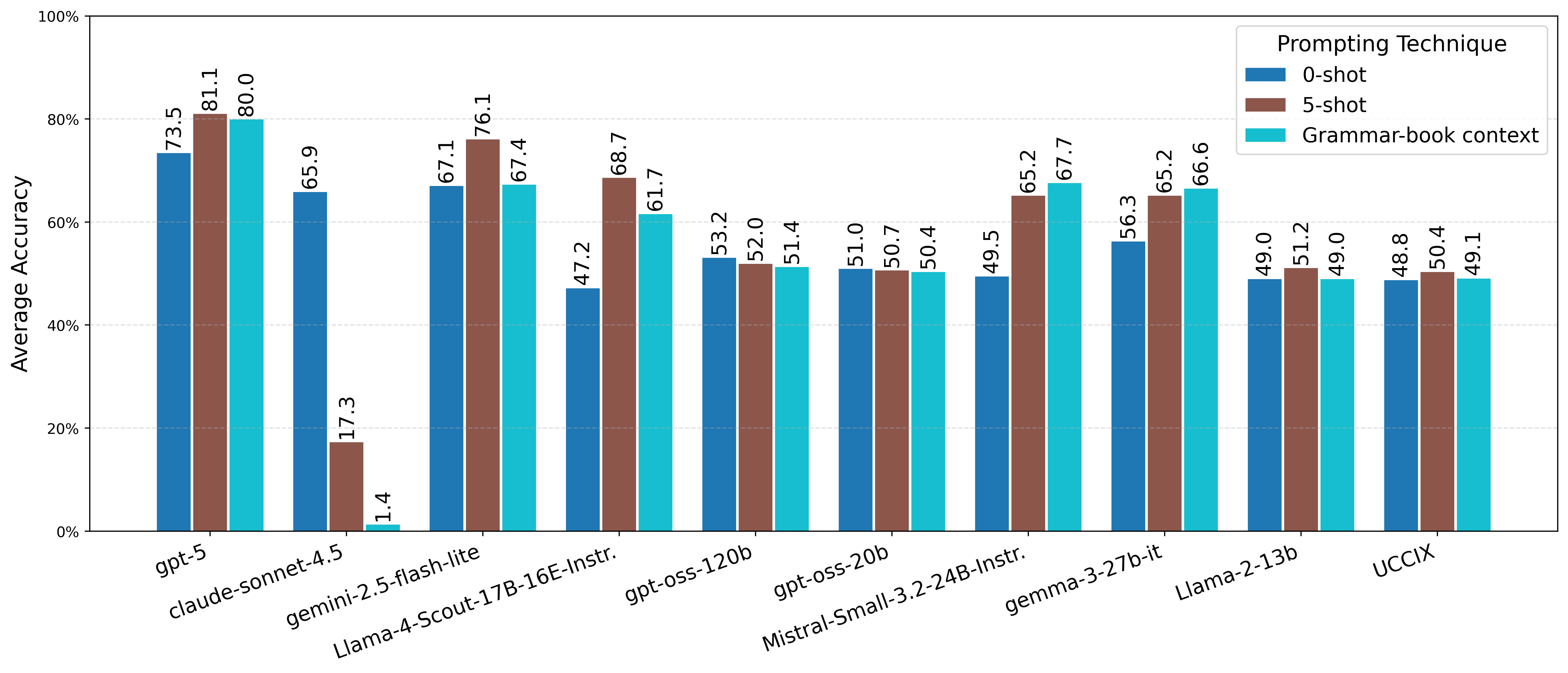}
    \caption{Accuracy by model (groups) and prompting technique (bars).}
    \label{fig:prompting}
\end{figure*}

\section{Results \& Discussion}
	\subsection{Performance Gap Between LLMs and Human Baseline}
		Figure~\ref{fig:zeroshot} presents the heatmaps of zero-shot prompting accuracy across models and categories. A clear gap emerges between human performance and all evaluated LLMs. Human participants achieved an average accuracy of 90.1\%, whereas the top-performing models, gpt-5 and gemini-2.5-flash-lite, reached only 73.5\% and 67.1\%, respectively. Furthermore, humans consistently outperform all models across all categories, ranking first in all 11 diverse linguistic phenomena.

		This trend contrasts with the original English BLiMP benchmark~\cite{warstadt2020blimp} - intentionally designed to challenge humans and language models alike. In our case, our Irish dataset appears easy for humans but difficult for LLMs, underscoring the disparity in LLM capabilities across extremely low-resource languages. These results highlight the imbalance of multilingual competence and the need for further research on grammar understanding in low-resource linguistic contexts.
	
		\subsubsection{Closed-source vs. Open-source Models}
			Figure~\ref{fig:zeroshot} also reveals a consistent gap between closed-source (API-only) and open-weight models. gpt-5, claude-sonnet-4.5, and gemini-2.5-flash-lite outperform open models by a wide margin (68.8\% compared to 50.7\%, averaged across models of the same type), illustrating the performance disparity between proprietary and open models.

			Interestingly, gemma-3-27b-it (27B parameters) outperforms models much larger in size (e.g., Llama-4-Scout-17B-16E-Instr. - 109B parameters and gpt-oss-120b - 120B parameters) in overall accuracy. However, this advantage is inconsistent across categories: it underperforms by up to 20.0\% on ``Adjectives and Comparison'' compared to Llama-4-Scout-17B-16E-Instr., but surpasses larger models by over 10 points on ``Clause Structure and Word Order''. Nevertheless, all open-source models perform near random chance ($\approx 50.0\%$), indicating that true grammatical understanding of Irish remains limited across current open architectures.

		\subsubsection{Effects of Language-specific Fine-tuning}
			Contrary to expectations, language-specific fine-tuning does not lead to significant gains. The Irish-adapted UCCIX model slightly underperforms its base model (Llama-2-13B) by 0.2\% average accuracy. This suggests fine-tuning alone may be insufficient when the available Irish corpus is small, leading to overfitting or limited generalisation, and more efforts are needed for effective multilingual transfer for extremely low-resource languages.

	\subsection{Performance Variation Across Linguistic Phenomena}
		We next examine whether certain linguistic phenomena are consistently more challenging for LLMs, if what LLMs found difficult are similar to what humans found difficult. Figure~\ref{fig:correlation} displays the Pearson correlation between human and model accuracies across categories. All LLMs show weak correlation with human performance ($-0.7 < r < 0.7$), with the strongest being claude-sonnet-4.5 ($r = 0.47$). This weak alignment indicates that the types of errors made by models differ from those made by humans, i.e., models struggle with aspects of Irish grammar that humans find trivial.

		Moreover, fewer than half of the ten evaluated LLMs exhibit strong inter-model correlation, suggesting that their internal linguistic representations diverge substantially. This reinforces that current LLMs have not yet developed a human-like representation of Irish grammar.

        \begin{table}[h]
        \centering
        \begin{tabular}{lc}
        \toprule
        \textbf{Model} & \textbf{std.} \\
        \midrule
        gpt-5 & 8.41 \\
        claude-sonnet-4.5 & 12.24 \\
        gemini-2.5-flash-lite & 8.55 \\
        Llama-4-Scout-17B-16E-Instr. & 5.97 \\
        gpt-oss-120b & 6.51 \\
        gpt-oss-20b & 5.53 \\
        Mistral-Small-3.2-24B-Instr. & 5.43 \\
        gemma-3-27b-it & 8.16 \\
        Llama-2-13b & 7.69 \\
        UCCIX & 7.61 \\
        \hline
        \textbf{Human Baseline} & \textbf{4.29} \\
        \bottomrule
        \end{tabular}
        \caption{Standard deviation across categories.}
        \label{tab:std}
        \end{table}

		Table~\ref{tab:std} reports the standard deviation of accuracy across phenomena. Human performance shows low variance ($\sigma = 4.29$), implying uniform difficulty. In contrast, LLMs exhibit much higher variance (up to 12.24 for claude-sonnet-4.5), revealing instability in syntactic and morphological understanding across grammatical contexts.

	\subsection{Prompting Technique Comparison}
		Figure \ref{fig:prompting} summarises the accuracy under different prompting strategies. We observe a failure mode for claude-sonnet-4.5 when additional context is provided: instead of producing letter label (``A'' or ``B'') required by the exact-match metric, the model generates extended explanations, leading to failures, potentially due to more attentions are given to the context rather than task requirement.

		Overall, both few-shot and grammar-context prompts improve performance for most models, though neither consistently dominates. However, this is not universal, as for some families of model (e.g., gpt-oss and Llama-2), no clear differences are observed over 0-shot prompting. Nevertheless, Llama-4 improves from 47.2\% (0-shot) to 68.7\% (5-shot), more than 20.0\% increase, showcasing strong in-context learning. The best absolute performance is achieved by gpt-5 under 5-shot prompting (81.1\%), still below the human baseline (88.0\%), even though models receive more contextual information than human evaluators.
	
		These results indicate that in-context learning boosts surface accuracy but does not yet yield deep grammatical understanding, and LLMs appear to rely on pattern recognition rather than internalised linguistic rules.

\section{Conclusion}
    In this work, we contribute Irish-BLiMP, a novel dataset for the syntactic evaluation of LLMs and their linguistic understanding of the Irish language, an extremely low-resource language. Our experiments reveal a substantial performance gap: the benchmark is straightforward for humans (average accuracy 90.1\%), yet remains highly challenging for LLMs. State-of-the-art open-source models perform near the random baseline, while the strongest closed-source model, gpt-5, achieves only 73.5\% accuracy. This contrast indicates that current models rely primarily on surface-level pattern recognition rather than internalised grammatical competence in Irish. Irish-BLiMP serves as a valuable benchmark for advancing research on linguistic understanding in low-resource languages. Future research can focus on expanding coverage of the dataset, e.g., to dialectal variations of Irish.

\section{Limitations}
    This paper focuses specifically on the Irish language, an endangered language. In principle, our framework for creating and evaluating the grammatical competence of LLMs can be extended to other languages as well. Furthermore, our dataset, while covering diverse linguistic phenomena, focuses on written standard Irish, which does not fully capture dialectal diversity. Future work should expand the dataset to include dialect-specific constructions and phonologically conditioned variation.

\section{Ethics statement}
    Our dataset is constructed in alignment with the linguistic features and examples from publicly available sources, ensuring that no copyrighted or sensitive textual material is included. Our work contributes to language technologies that support the digitalisation and preservation of endangered languages, with a particular focus on Irish. We further acknowledge that LLMs trained predominantly on high-resource languages can exhibit systemic linguistic biases. Irish-BLiMP is intended to highlight such disparities and to promote further multilingual model development.

\nocite{*}
\section{Bibliographical References}\label{sec:reference}

\bibliographystyle{lrec2026-natbib}
\bibliography{lrec2026-example}


\end{document}